\documentclass[singlecolumn]{openhelix}
\usepackage[most]{tcolorbox}

\usepackage{amsmath}
\usepackage{amssymb}
\usepackage{tabularx}
\usepackage{float}
\usepackage{utfsym}
\usepackage{array}
\usepackage{booktabs}
\usepackage{makecell}
\usepackage{xspace}

\newcommand{\huggingface}{\raisebox{-1.5pt}{\includegraphics[height=1.05em]{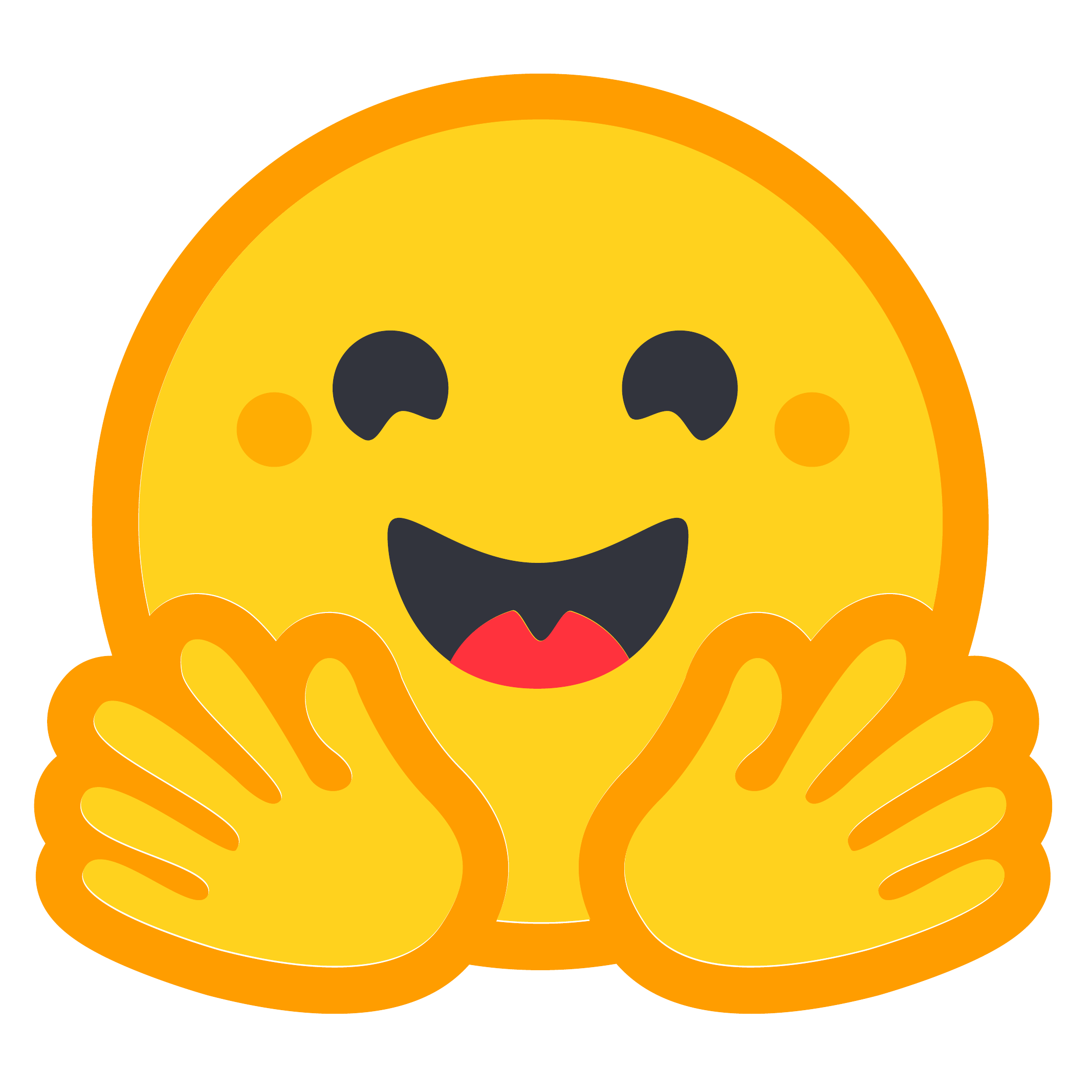}}\xspace}

\newcommand{\github}{\raisebox{-1.5pt}{\includegraphics[height=1.05em]{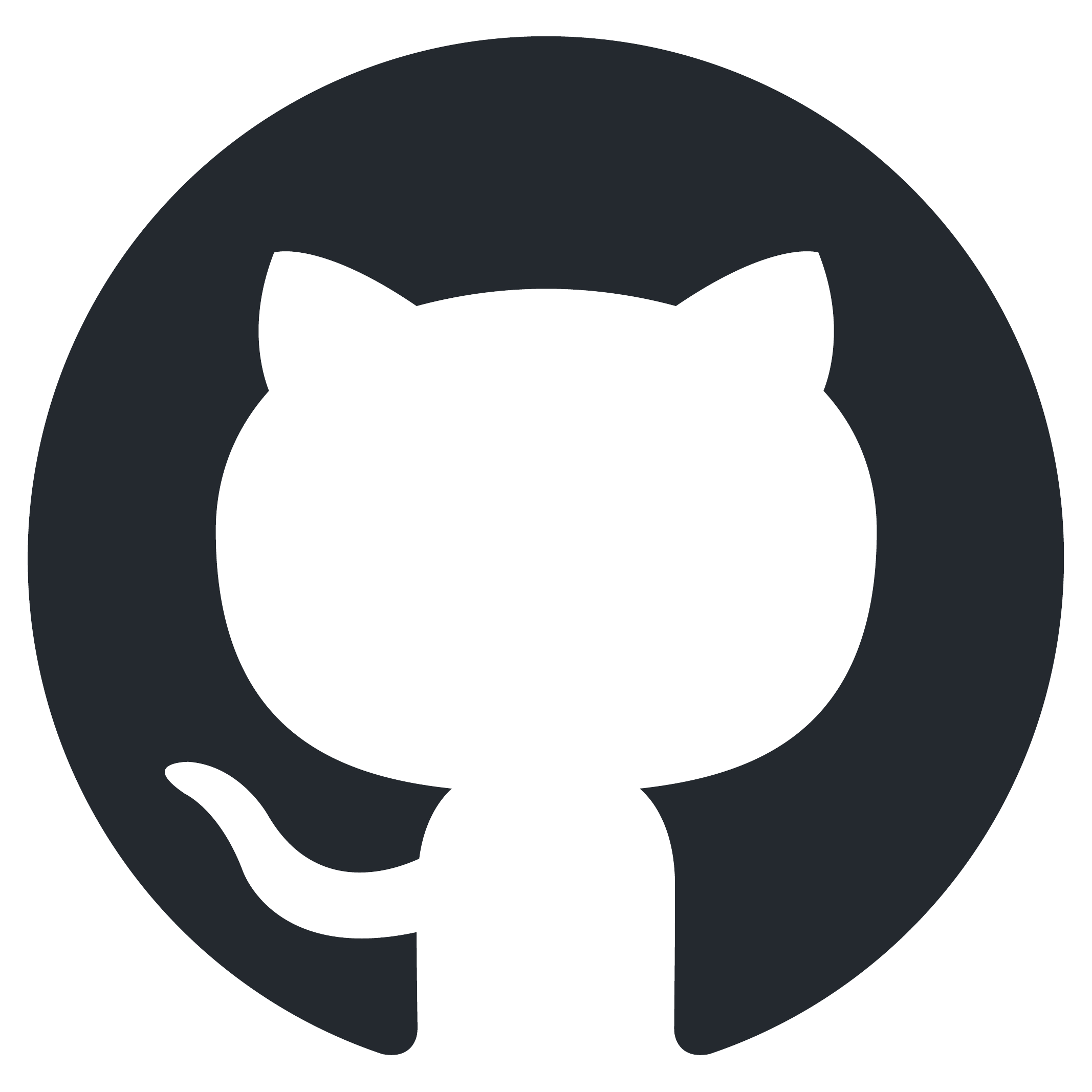}}\xspace}
\newcommand{\website}{\raisebox{-1.5pt}{\includegraphics[height=1.05em]{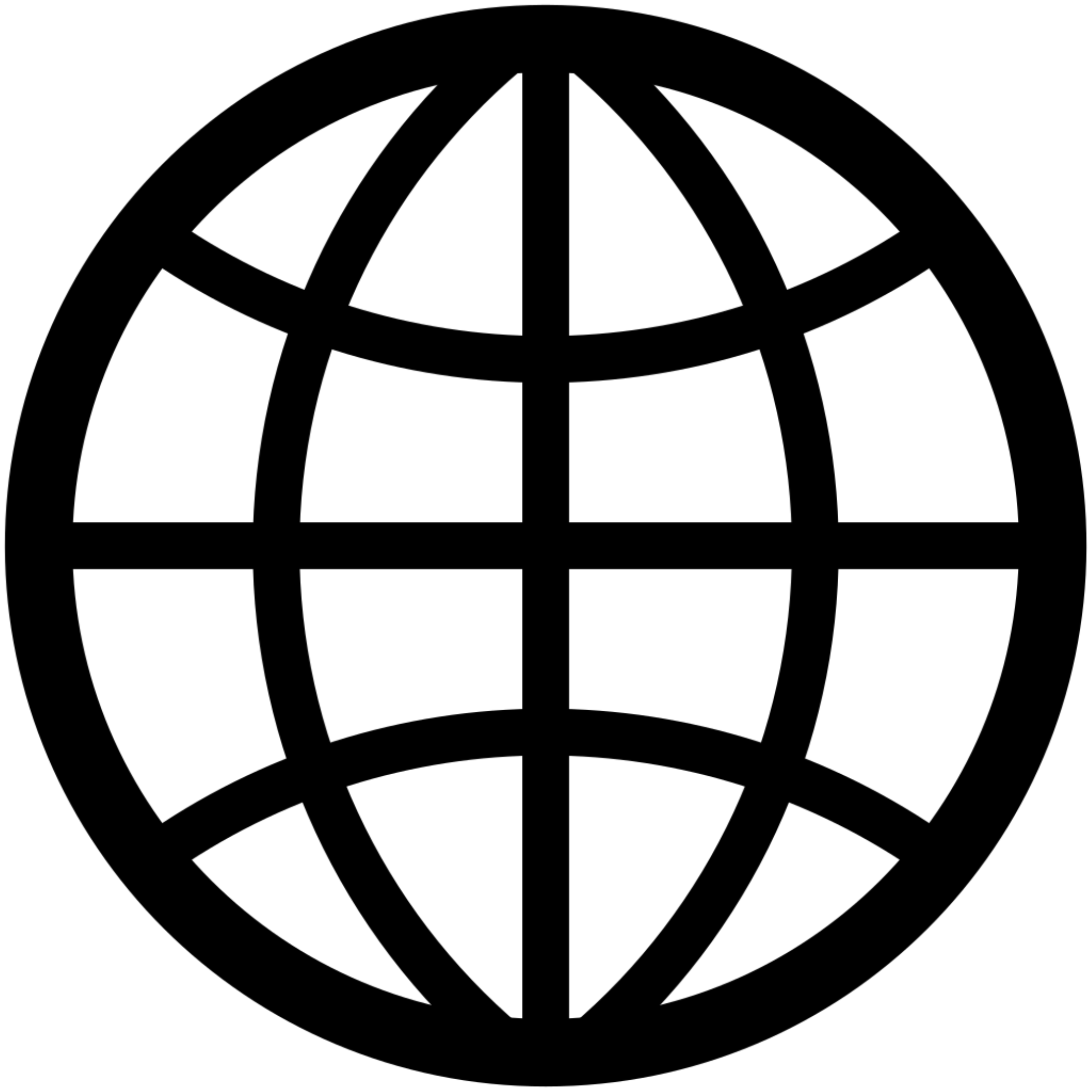}}\xspace}

\usepackage[table]{xcolor}
\definecolor{newyellow}{HTML}{FFD94D}
\definecolor{newgrey}{HTML}{7F7F7F}
\definecolor{newpink}{HTML}{FBCDF4}
\newcommand{\method}{\texttt{FRAPPE}}

\title{\texorpdfstring{\raisebox{-0.1\height}{\includegraphics[height=3em]{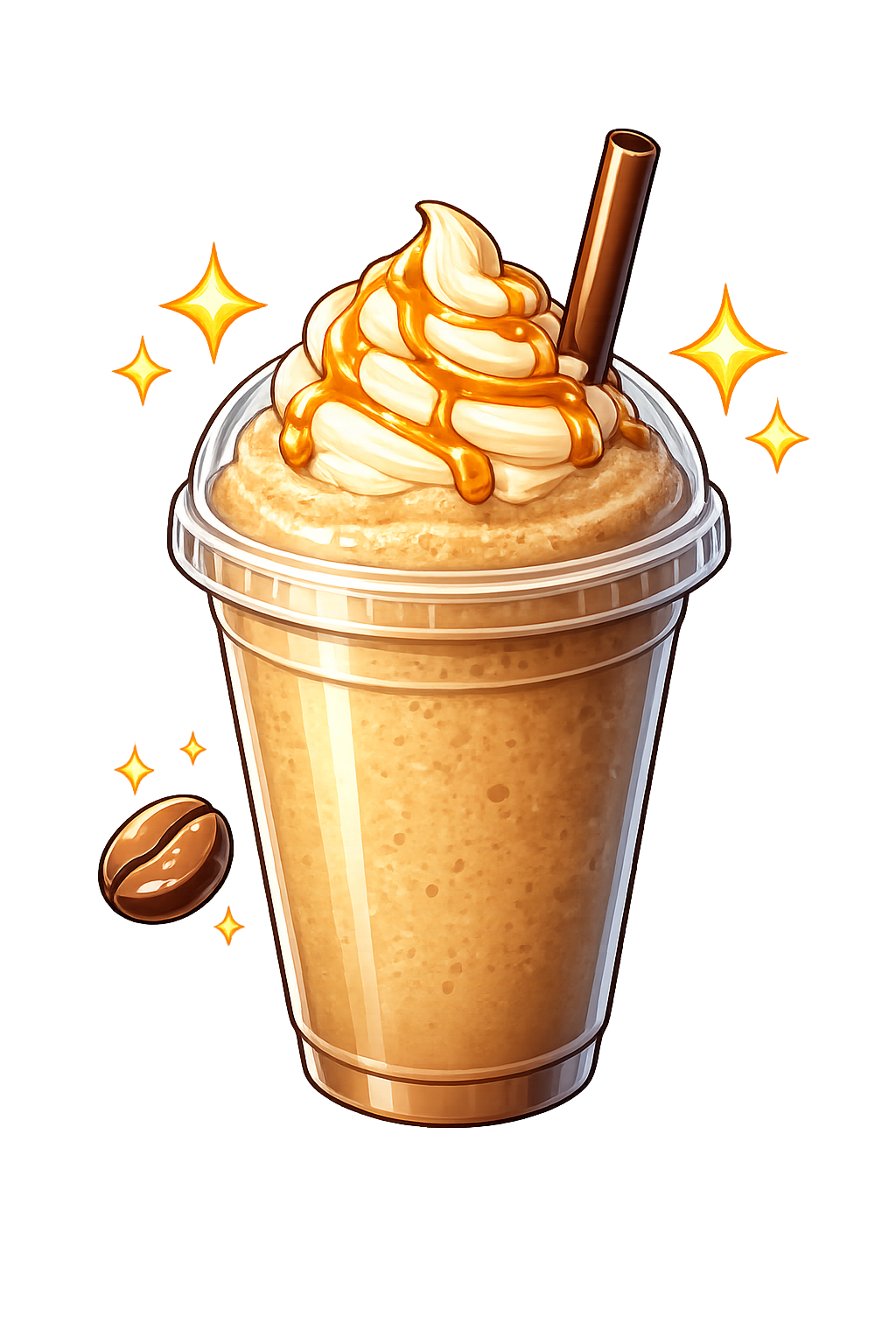}}}{}FRAPPE: Infusing World Modeling \\ into Generalist Policies via Multiple Future Representation Alignment}

\author[1,2\dagger\ddagger]{Han Zhao}
\author[3,4\dagger]{Jingbo Wang}
\author[3\dagger]{Wenxuan Song}
\author[5]{Shuai Chen}
\author[2]{Yang Liu}
\author[6]{Yan Wang}
\author[3*]{Haoang Li}
\author[2*]{Donglin Wang}

\affiliation[1]{Zhejiang University}
\affiliation[2]{Westlake University}
\affiliation[3]{HKUST (GZ)}
\affiliation[4]{South China University of Technology}
\affiliation[5]{ShanghaiTech University}
\affiliation[6]{Tsinghua University}

\contribution[\dagger]{Equal Contribution}
\contribution[\ddagger]{Project Lead}
\contribution[*]{Corresponding Authors}

\abstract{
Enabling VLA models to predict environmental dynamics, known as world modeling, has been recognized as essential for improving robotic reasoning and generalization. 
However, current approaches face two main issues: 1. The training objective forces models to over-emphasize pixel-level reconstruction, which constrains semantic learning and generalization 2. Reliance on predicted future observations during inference often leads to error accumulation.
To address these challenges, we introduce \textbf{F}uture \textbf{R}epresentation \textbf{A}lignment via \textbf{P}arallel \textbf{P}rogressive \textbf{E}xpansion (\method). 
Our method adopts a two-stage fine-tuning strategy: In the mid-training phase, the model learns to predict the latent representations of future observations; In the post-training phase, we expand the computational workload in parallel and align the representation simultaneously with multiple different visual foundation models.
By significantly improving fine-tuning efficiency and reducing dependence on action‑annotated data, \method~provides a scalable and data‑efficient pathway to enhance world‑awareness in generalist robotic policies.
Experiments on the RoboTwin benchmark and real‑world tasks demonstrate that \method~outperforms state-of-the-art approaches and shows strong generalization in long‑horizon and unseen scenarios. 
}

\correspondence{%
\begin{tabular}[t]{@{}l@{\ }l@{}}
Han Zhao at & \email{zhaohan34@westlake.edu.cn}
\end{tabular}%
}

\metadata[\website Project Website]{\href{https://h-zhao1997.github.io/frappe}{\texttt{https://h-zhao1997.github.io/frappe}}}

\metadata[\huggingface Model]{\href{https://huggingface.co/collections/hhhJB/frappe}{\texttt{https://huggingface.co/collections/hhhJB/frappe}}}

\metadata[\github Code]{\href{https://github.com/Jbo-Wang/frappe}{\texttt{https://github.com/Jbo-Wang/frappe}}}

\begin{document}

\maketitle

\abstract{
Enabling VLA models to predict environmental dynamics, known as world modeling, has been recognized as essential for improving robotic reasoning and generalization. 
However, current approaches face two main issues: 1. The training objective forces models to over-emphasize pixel-level reconstruction, which constrains semantic learning and generalization 2. Reliance on predicted future observations during inference often leads to error accumulation.
To address these challenges, we introduce \textbf{F}uture \textbf{R}epresentation \textbf{A}lignment via \textbf{P}arallel \textbf{P}rogressive \textbf{E}xpansion (\method). 
Our method adopts a two-stage fine-tuning strategy: In the mid-training phase, the model learns to predict the latent representations of future observations; In the post-training phase, we expand the computational workload in parallel and align the representation simultaneously with multiple different visual foundation models.
By significantly improving fine-tuning efficiency and reducing dependence on action‑annotated data, \method~provides a scalable and data‑efficient pathway to enhance world‑awareness in generalist robotic policies.
Experiments on the RoboTwin benchmark and real‑world tasks demonstrate that \method~outperforms state-of-the-art approaches and shows strong generalization in long‑horizon and unseen scenarios. 
}
\section{Introduction}
\begin{figure*}[t]  
    \centering
    \includegraphics[width=0.99\linewidth]{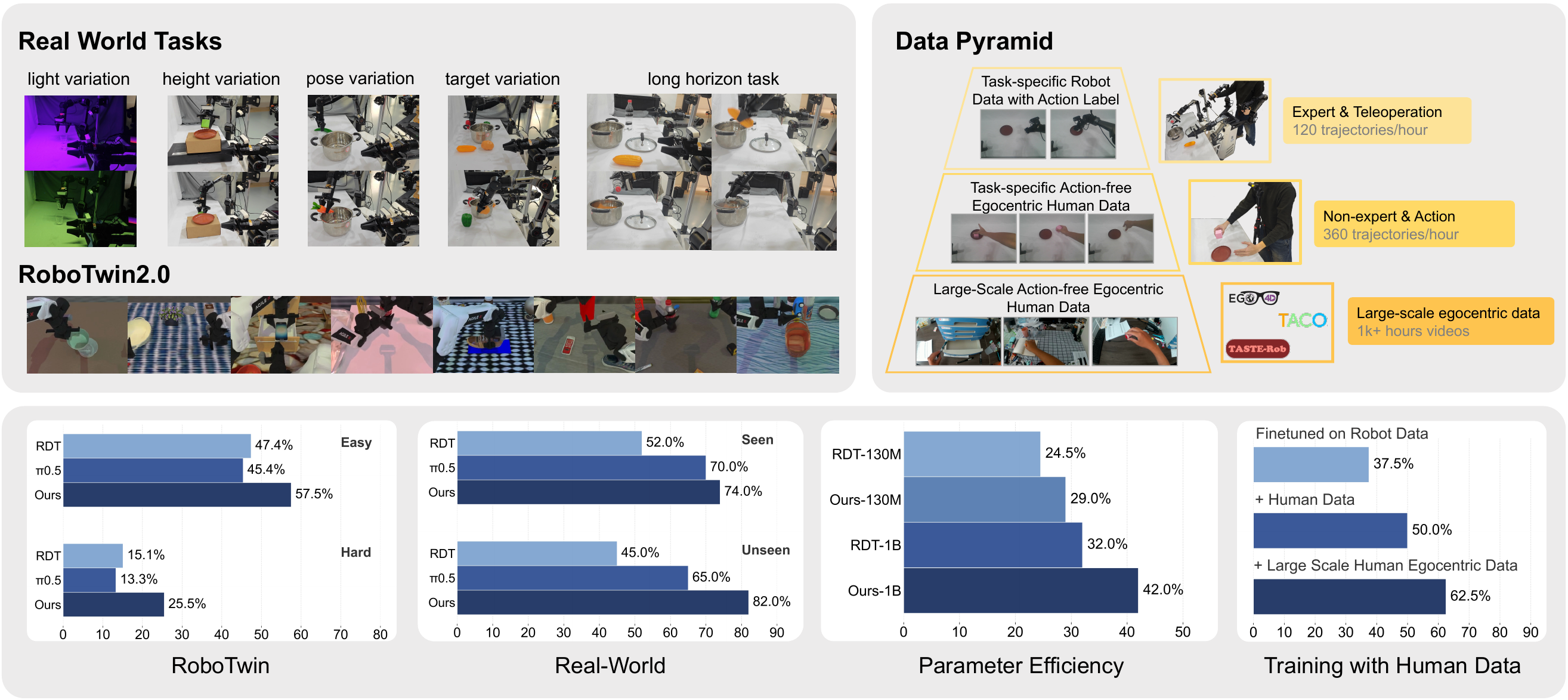}
    \caption{We demonstrate that \method~significantly outperforms the state-of-the-art models in both simulated and real-world complex scenarios, and it can effectively leverage data from different levels of the training data pyramid.}
    \label{fig:teaser}
\end{figure*}

Generalist robotic policies, often built on diffusion generative models~\citep{dp, dp3, rdt, rdt2}, have achieved significant advances in manipulation and navigation tasks, as they can effectively model multi-modal action distributions. Recent approaches~\citep{pad, uva, uwm, vpp, genie, dust, unipi} further extend diffusion-based policies with auxiliary tasks in the world-model manner (\textit{i.e.}, predicting future images). 
These methods incorporate observation supervision to help the model capture scene dynamics, thereby mitigating overfitting to action histories and enhancing robustness to visual perturbations.

The above works have primarily focused on designing improved architectures to integrate world modeling objectives into policies, while neglecting the investigation of crucial finetuning and inference recipes for deploying these policies. The oversight makes these works reveal two significant limitations: 
1. For explicit world modeling policies, generating pixels causes the model to allocate substantial computational resources to fitting redundant pixels, rather than focusing on task-relevant object visual information~\citep{reconvla}. This results in poor generative quality in out-of-distribution (OOD) scenarios. Also, during inference, relying on explicitly predicted observations for action generation may also lead to error accumulation~\citep{unipi, dreamitate}.
2. Some prior work has attempted to embed implicit world models/knowledge in the network via representation alignment~\citep{flare, vpp, sf}. However, representations learned from a single visual task inherently carry inductive biases, making them not necessarily suitable for all tasks.

To address these limitations, we propose \textbf{F}uture \textbf{R}epresentation \textbf{A}lignment via \textbf{P}arallel \textbf{P}rogressive \textbf{E}xpansion (\textbf{\method}).
Our philosophy is to enhance the model's performance through scaling the computational workload and the implicit grounding of future observations during both training and inference simultaneously. This is realized by a variation of \textit{parallel scaling}~\citep{parallelscaling}. We expand the input streams and training multiple expert networks.
The different expert networks share the same frozen backbone, while each has its own set of learnable tokens and Low-Rank Adaptation (LoRA) modules~\citep{lora}. The actions are then aggregated by a learnable router to obtain the final action. The entire model can also be viewed as a mixture-of-experts model, which is also widely applied in the architectural design of VLA models~\citep{germ, more}.
During \textbf{training}, in addition to the action supervision objective, we utilize diverse pretrained visual foundation models (VFMs) to encode future observations as alignment targets. 
Each output future prefix is supervised by the embedding of future observations produced by the corresponding VFM.
With the training objective and the prefix-and-LoRA architecture, the finetuning recipe achieves strong performance with minimal trainable parameters.
During \textbf{inference}, we retain the same computational graph without incorporating VFMs for supervision.
In this way, the model benefits from multiple alignment objectives and multi-stream forward to obtain the scaling advantage.

\begin{figure*}[t]  
    \centering
    \includegraphics[width=0.9\linewidth]{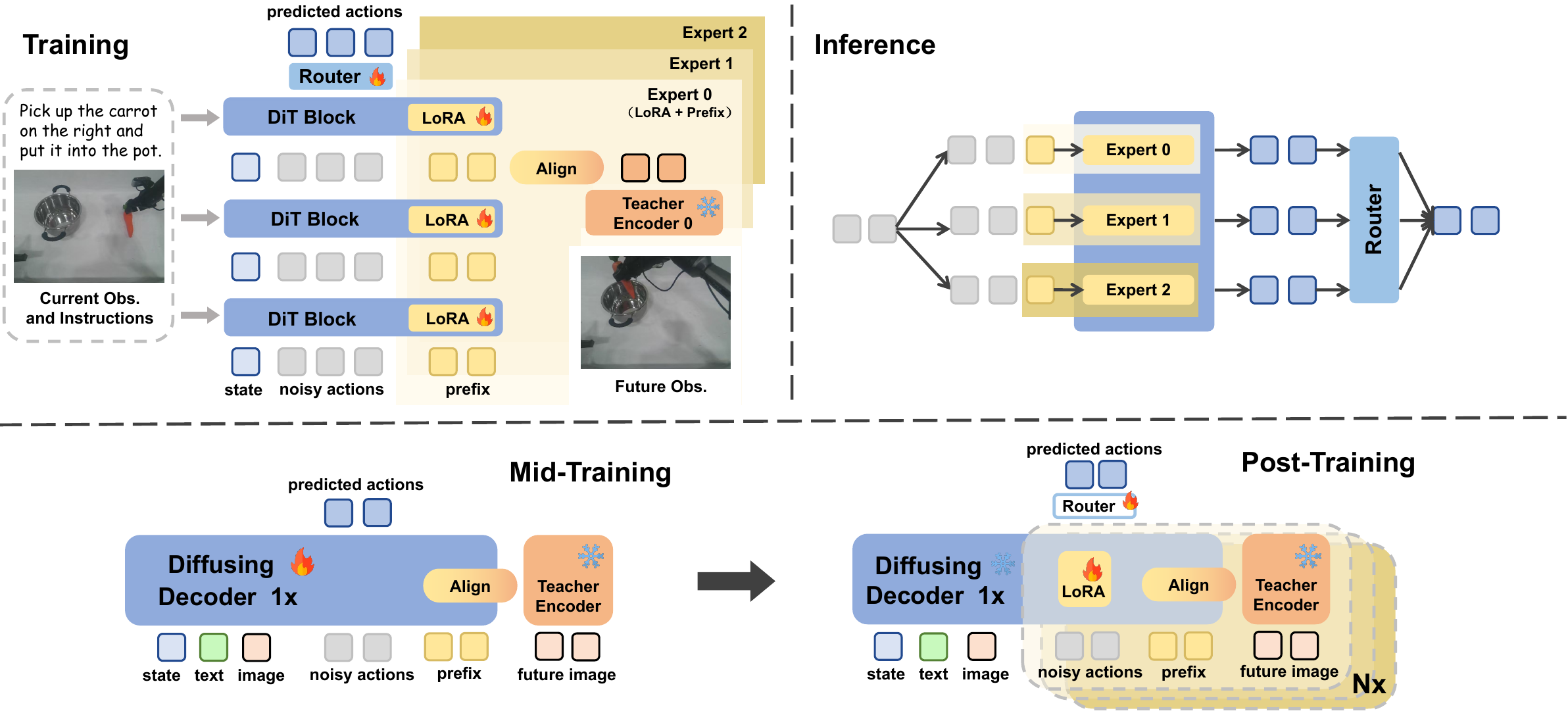} 
    \caption{\textbf{Overview of training and inference.} During the training phase, the model progressively learns to align with the representation spaces of multiple visual foundation models simultaneously. The model is trained through a two-stage training process to extends to parallel processing of multiple input streams while aligning diverse visual representations. Similarly, parallel inference is implemented during the inference stage.}
    \label{fig:overview} 
\end{figure*}
In practice, we empirically observe that directly training a foundation model for parallel scaling and world-modeling capabilities is notably challenging, as both the model architecture and the training objectives undergo significant shifts compared to those in the pretraining stage.
We thus improve it by a two-stage approach: The \textbf{mid-training} stage aligns our single model with a teacher encoder distilled from our multiple VFMs. During this process, the model is full-parameter finetuned to adapt our model to the world-modeling objective.
The parallel training is applied during the \textbf{post-training} stage to achieve scaling in computing and assign each expert a distinct visual representation from a different encoder for supervision.
Beyond the performance gains from model architecture and training objectives, our world-modeling training process can significantly benefit from the incorporation of action-free data.
\textbf{In small-scale fine-tuning scenarios, }our model can substantially reduce the dependence on expert teleoperation data collection, replacing it with more efficiently collected human hand‑manipulation data. \textbf{The large‑scale internet egocentric data can also scale up training}, the training of world‑modeling capabilities can be extended into a form of continual pre‑training on conventional VLA base models.

In summary, the contributions of our method are as follows:

\begin{itemize}
    \item We propose a novel training paradigm that enhances the implicit world modeling capacity of VLA models. \method~builds a robust implicit world model by aligning the model's features with multiple visual latent representations. By scaling up computation, \method~enhances generative capabilities while avoiding the inductive biases inherent in single-representation learning.

    \item We design a two-stage training strategy that allows the model to progressively expand its capabilities. This enables efficient learning and avoids slow convergence that occurs when parallel scaling is applied directly.

    \item Our approach can leverage human video demonstrations without action annotations for training, allowing VLA models to scale training data at lower cost compared with imitation learning that relies solely on action-labeled data.

    \item In both simulation and real-world settings, our approach surpasses the state-of-the-art (SOTA) models. Notably, when teleoperation data is extremely limited, our method can learn from human-executed videos and improve overall performance by~10-15\% compared with a teleoperation-only baseline.
    
\end{itemize}
\section{Related Work}
\paragraph{Diffusion-based Policies}
Diffusion models enable policies to capture multi-modal action distributions and ensure temporal consistency. Diffusion Policy~\citep{dp} introduced this conditional denoising formulation, which DP3~\citep{dp3} and iDP3~\citep{idp3} advanced by integrating 3D representations to achieve superior spatial generalization and robust humanoid control. Recently, RDT~\citep{rdt} scaled this paradigm to a foundation model level using a diffusion transformer (DiT) trained on massive heterogeneous datasets. Building upon the original RDT, RDT2~\citep{rdt2} further scales up the parameter capacity and incorporates a specialized action tokenizer alongside a flow-matching loss. 
In contrast, our method provides an accessible approach to further enhance capabilities through future image prediction.
\paragraph{Generative Models Unifying Vision and Action}
Recent approaches have sought to unify video generation and action prediction within diffusion frameworks. UVA~\citep{uva} learns a joint video-action latent representation but employs decoupled lightweight diffusion heads for decoding. PAD~\citep{pad} and UD-VLA~\citep{udvla} unifies the process by encoding modalities separately but performing joint denoising to generate future frames and actions. 
Alternatively, rather than explicit generation, Genie Envisioner~\citep{genie} and VPP~\citep{vpp} leverage video diffusion models as encoders, extracting predictive visual representations to condition the downstream policy. Similarly, FLARE~\citep{flare} integrates implicit world modeling into policy learning by aligning the introduced future embeddings with the encoded future observations.
In contrast, our method optimizes the model by parallel scaling the implicit world modeling process.

\paragraph{Robot Learning from Human Egocentric Videos}

Prior research in robot learning from human videos has primarily extracted implicit representations, such as object affordances~\citep{hrp, videodex, affordances} or visual trajectories~\citep{affordances,track2act}.
While various works employ pose estimators~\citep{videodex, zhu2024vision, mimicplay, ye2023learning} to translate human motions into robotic behaviors, Being-H0~\citep{beingh0} and EgoVLA~\citep{egovla} prioritize high-fidelity 3D hand modeling to achieve the precision required for dexterous manipulation.
In contrast, our approach obviates the need for explicit pose estimators, instead leveraging a straightforward prefix-finetuning strategy to distill salient information from egocentric videos.
Another research paradigm focuses on extracting latent actions from the temporal transitions between current and future frames to guide downstream learning~\citep{videoworld,ye2024latent, bruce2024genie,dreamitate,bu2025agibot}.
Beyond these engineered mappings, Kareer et al.~\citep{emergence} demonstrate that human-to-robot transfer can arise as an emergent property from diverse co-training across embodiment-agnostic datasets. 

In contrast, our framework employs future observations as an explicit grounding signal rather than as ambiguous latent features. By doing so, we endow the model with robust dynamic modeling capabilities, which significantly strengthen the efficacy of policy learning.
\section{Method}

In this section, we first introduce our future representation alignment strategy and its implementation during the mid-training phase. Subsequently, we present Mixture-of-Prefix-and-LoRA (MiPA) and its realization in the post-training stage. 
Finally, we elaborate on the inference-phase parallel scaling.
\subsection{Preliminaries: Robotic Diffusion Transformer} 
Robotic Diffusion Transformer (RDT)~\citep{rdt} is a pioneering foundation model for bimanual manipulation. RDT models the conditional distribution of future action sequences, $p_\theta(\textbf{a}_t | \textbf{o}_t, l)$. It trains a denoising network, $f_\theta$, to predict a clean action chunk $\textbf{a}_t$ from a noisy version $\tilde{\textbf{a}}_t$ conditioned on the observation $\textbf{o}_t$ and language instruction $l$. The training objective is to minimize the mean-squared error (MSE):
\begin{equation}
    \mathcal{L}_{\text{action}} := \text{MSE} \left( \textbf{a}_t, f_\theta(l, \textbf{o}_t, \tilde{\textbf{a}}_t, k) \right),
\label{eq:rdt_loss}
\end{equation}
where $\tilde{\textbf{a}}_t = \sqrt{\bar{\alpha}_k}\textbf{a}_t + \sqrt{1 - \bar{\alpha}_k}\epsilon$ with noise $\epsilon \sim \mathcal{N}(0, \mathbf{I})$ and a diffusion timestep $k$. 

The network $f_\theta$ is implemented as a Diffusion Transformer (DiT). It takes low-dimensional state tokens (including $\tilde{\textbf{a}}_t$ and proprioception $\textbf{z}_t$) as its input sequence. This sequence is then processed through a series of cross-attention layers conditioned on high-dimensional visual tokens $\{x_i^V\}$ (from a SigLIP encoder~\citep{siglip}) and language tokens $\{x_j^L\}$ (from a T5 encoder~\citep{T5}), Finally, the output of the backbone is projected back to the physical action space by a MLP decoder to produce the action chunk $\hat{A}_t$. 

\subsection{Future Representation Alignment via Parallel Progressive Expansion}
\textbf{Overview. }As illustrated in \Cref{fig:overview}, the overall philosophy of our method is to construct a world-modeling task during the training process.
An inductive idea is to make the model generate future states along with actions.
While the visual information is used as the condition for RDT, and can not be supervised directly, the process can be realized by adding a set of noisy prefixes as input, and the model is expected to predict future states from them. 
Specifically, for the input of the RDT includes proprioception $z_t$, noisy actions $\tilde{a}_t$, control frequency $c$, and diffusion timestep $k$, we introduce a set of learnable tokens $\mathbf{p} \in \mathbb{R}^{n \times d}$ as the future prefix to be concatenated with the input. 
To avoid pixel redundancy, we leverage pretrained vision foundation models as the teacher visual encoder $\Phi$ to provide supervising signals $\mathbf{e}_{t+h}=\Phi(o_{t+h}), \mathbf{e} \in \mathbb{R}^{n \times d_\Phi}$, where $h$ denotes the future horizon, $n$ and $d_\Phi$ is the length and dimension of output. 
The denosing process is formulated as
\begin{equation}
    \textbf{a}_t,\textbf{p}_t=f_\theta(l, \textbf{o}_t, \tilde{\textbf{a}}_t, k).
\label{eq:denosing_process}
\end{equation}
The alignment loss for a VFM $\Phi$ is formulated as
\begin{equation}
    \mathcal{L}_\Phi = \textnormal{cos}(\mathbf{p}_t, \text{sg}(\mathbf{e})),
\label{eq:align_loss_S}
\end{equation}
where $\textnormal{cos}[\cdot, \cdot]$ denotes cosine similarity and $\text{sg}(\cdot)$ is the stop-gradient operator, ensuring that gradients do not flow back into the teacher model.
Thus, the training objective is a combination of action loss and alignment loss.
This process makes the model learn to predict the future states and model the dynamics between the actions and states.

\paragraph{Parallel Scaling}

To leverage the world knowledge within diverse VFMs and conduct parallel scaling, we propose a prefix-and-LoRA tuning paradigm to align with multiple future visual representations.
Specifically, we construct multiple sets of future prefixes and LoRAs in a shared RDT backbone, and each set is activated during alignment with a dedicated teacher encoder.
To reduce the memory footprint during fine-tuning, only the prefixes and LoRAs are trainable in the framework.
Considering that all MiPAs are supervised, the loss function should be re-formulated as:
\begin{equation}
    \mathcal{L}_{\text{align}} = (\sum_{i=1}^M \mathcal{L}_{\Phi_i}),
\label{eq:align_loss}
\end{equation}
where $M$ is the number of teacher encoders, and $\mathcal{L}_{\Phi_i}$ denotes the $i$-th teacher encoder. In this paper, we set $M=3$, and the selected teacher encoders are CLIP~\citep{clip} (400M), DINOv2 ~\citep{dinov2} (142M), and ViT (300M)~\citep{vit}.

To aggregate the outputs of these parallel experts, we introduce a lightweight router network. The router computes a set of gating weights, $\textbf{w} = \{w_i\}_{i=1}^{M}$, for the $M$ experts, such that $\sum_{i=1}^{M} w_i = 1$. Let the latent action representation from the $i$-th expert be $z_j$. The final executable action chunk, $\textbf{a}_t$, is then produced by passing the weighted sum of these latent representations through a shared action head, $\text{MLP}(\cdot)$:
\begin{equation}
    \textbf{a}_t = \text{MLP}\left( \sum_{i=1}^{M} w_i \cdot z_i \right).
\label{eq:moe_full_process}
\end{equation}

\paragraph{Load Balance during Output Aggregation} 
During training, we observed a mode collapse phenomenon, where a single stream would dominate the learning process, preventing others from updating \citep{switch,loadbalance,labelsmoothing}. To mitigate this issue and encourage balanced expert utilization, we incorporate a load-balancing loss into our training strategy:

\begin{equation}
    \mathcal{L}_{\text{balance}} = \frac{1}{B} \sum_{j=1}^{B} \left( \log \sum_{i=1}^{M} e^{\textbf{g}_{i,j}} \right)^2,
\label{eq:load_balance}
\end{equation}
where $\textbf{g}_{i,j}$ is the logit for the $j$-th token assigned to the $i$-th expert, and $B$ is the number of tokens in a batch. This encourages the router to produce logits with similar magnitudes across all experts.

Additionally, to prevent the router from assigning near-zero weights to certain experts in the early stages of training, we apply label smoothing directly to the final gating weights $w$. For each expert $i$, its weight $w_i$ is adjusted as follows:
\begin{equation}
    w'_i = w_i \cdot (1 - \epsilon) + \frac{\epsilon}{M},
\label{eq:label_smoothing}
\end{equation}
where $w'_i$ is the smoothed weight, and $\epsilon$ is a small hyperparameter ($0.1$ in this paper). This ensures that every expert receives a minimum, non-zero weight, thereby guaranteeing that all experts are updated. 

The final training objective for our framework is formulated as a composition of \Cref{eq:rdt_loss}, \Cref{eq:align_loss}, and \Cref{eq:load_balance}:
\begin{equation}
    \mathcal{L}_{\text{total}} = \mathcal{L}_{\text{action}} + \lambda_1 \mathcal{L}_{\text{align}} + \lambda_2 \mathcal{L}_{\text{balance}},
\label{eq:total_mipa_loss}
\end{equation}
with weight factors $\lambda_1$ and $\lambda_2$ that balance the contributions of the alignment and load-balancing objectives, respectively.

\paragraph{Mid-Training}
We find that the finetuning process can be further optimized through a mid-training process.
Before post-training, the mid-training stage employs traditional single-stream training to align the newly added future prefix with the teacher encoder.
To ensure consistency between the two stages and adapt single-stream alignment, we utilize a variant of Theia~\citep{theia}, which distills a tiny VFM from multiple pretrained VFMs to ensure comprehensive robotic capabilities. In this paper, we use the 86M encoder distilled from the three encoders used in the post-training phase.
This process adapts our model to the world-modeling objective, thereby obtaining a strong initialization for post-training.

\section{Simulation Experiments}
We choose to first conduct fair, uniform, and reproducible comparisons in simulation to demonstrate that our fine-tuning scheme exhibits significant advantages in performance (\Cref{subsec:comparison_sota}), training efficiency (\Cref{subsec:comparison_paradigm}), and efficiency across models of different parameter scales (\Cref{subsec:comparison_scale}). The following settings are consistent across all experiments in this section:
\paragraph{Simulation Environment}
We evaluate our method on RoboTwin \citep{robotwin}. 
RoboTwin is a real-to-sim bimanual benchmark.
It contains an \textit{Easy} setting with in-domain layout and a \textit{Hard} setting with domain randomization, including scene clutter, diverse background textures, lighting variation, and varied tabletop heights.
To comprehensively evaluate the performance, all simulation experiments are conducted on 8 diverse tasks under both \textit{Easy} and \textit{Hard} settings.
\paragraph{Training Details}

Our model is fine-tuned starting from the official RDT-1B pre-trained weights. For all experiments, the training dataset is restricted to 50 task-specific trajectories from the  Easy setting. The training process is conducted on two NVIDIA H100 GPUs for a total of 20,000 steps with a batch size of 32, partitioned into 15,000 mid-training steps and 5,000 post-training steps for all models using~\method. Accuracy is reported based on the average performance over 100 evaluation trials to ensure fair comparison.

\subsection{Comparison with State-of-the-Arts}
\label{subsec:comparison_sota}
Our primary baseline is the base RDT model finetuned using the original finetuning recipe.
However, for broader comparison, we also include several other baselines. The complete list of comparisons is as follows:

\begin{itemize}
    \item Diffusion Policy (\textbf{DP}~\citep{dp}): a train-from-scratch visuomotor policy, serving as a small-scale baseline model.
    
    \item Video Prediction Policy (\textbf{VPP}~\citep{vpp}): Training subsequent DP based on representations provided by a pre-trained video prediction diffusion model, serving as an implicit world model baseline for comparison.

    \item \textbf{RDT}~\citep{rdt}: The base model adopted for \method.

    \item $\mathbf{\pi_0}$~\citep{black2024pi_0}: the state-of-the-art method and its advanced version on RoboTwin Benchmark with a VLM and flow-matching head combined architecture.

    \item $\mathbf{\pi_{0.5}}$~\citep{intelligence2025pi_}: The successor to $\pi_0$, demonstrating stronger generalization capabilities.
\end{itemize}

\begin{table*}[t!]
\centering
\footnotesize
\caption{\textbf{Comparisons with state-of-the-art methods on RoboTwin 2.0 benchmark.} Please note that \textbf{Bold} and \underline{Underline} denote the best and the second-best performances among the methods.}
\label{tab:sim_tasks}

\begin{tabular*}{\linewidth}{@{\extracolsep{\fill}}l rr rr rr rr@{}}
\toprule
\multirow{2}{*}{\textbf{Method}} & \multicolumn{2}{c}{\textbf{Handover Mic}} & \multicolumn{2}{c}{\textbf{Pick Dual Bottles}} & \multicolumn{2}{c}{\textbf{Handover Block}} & \multicolumn{2}{c}{\textbf{Place Object Basket}} \\
\cmidrule(lr){2-3} \cmidrule(lr){4-5} \cmidrule(lr){6-7} \cmidrule(lr){8-9}
& Easy & Hard & Easy & Hard & Easy & Hard & Easy & Hard \\
\midrule
\textbf{DP} & 53.0\% & 0.0\% & 24.0\% & 0.0\% & 10.0\% & 0.0\% & 15.0\% & 0.0\% \\
\textbf{VPP} & \underline{91.0\%} & 7.0\% & 28.0\% & 6.0\% & 8.0\% & 0.0\% & 30.0\% & 0.0\% \\
\textbf{RDT} & 90.0\% & \underline{31.0\%} & 45.0\% & \underline{14.0\%} & \underline{45.0\%} & \underline{14.0\%} & 30.0\% & \underline{6.0\%} \\
\textbf{$\mathbf{\pi_0}$} & \textbf{98.0\%} & 13.0\% & \underline{57.0\%} & 12.0\% & \underline{45.0\%} & 8.0\% & 16.0\% & 2.0\% \\
\textbf{$\mathbf{\pi_{0.5}}$} & \textbf{98.0\%} & 13.0\% & 37.0\% & 13.0\% & 2.0\% & 0.0\% & \textbf{42.0\%} & 4.0\% \\ 
\textbf{Ours} & \textbf{98.0\%} & \textbf{45.0\%} & \textbf{67.0\%} & \textbf{28.0\%} & \textbf{50.0\%} & \textbf{18.0\%} & \underline{35.0\%} & \textbf{15.0\%} \\
\bottomrule
\end{tabular*}

\vspace{4mm}

\begin{tabular*}{\linewidth}{@{\extracolsep{\fill}}l rr rr rr rr | rr@{}}
\toprule
\multirow{2}{*}{\textbf{Method}} & \multicolumn{2}{c}{\textbf{Put Object Cabinet}} & \multicolumn{2}{c}{\textbf{Place Shoe}} & \multicolumn{2}{c}{\textbf{Stack Bowls Two}} & \multicolumn{2}{c|}{\textbf{Put Bottle Dustbin}} & \multicolumn{2}{c}{\textbf{Average}} \\
\cmidrule(lr){2-3} \cmidrule(lr){4-5} \cmidrule(lr){6-7} \cmidrule(lr){8-9} \cmidrule(lr){10-11}
& Easy & Hard & Easy & Hard & Easy & Hard & Easy & Hard & Easy & Hard \\
\midrule
\textbf{DP} & 42.0\% & 0.0\% & 23.0\% & 0.0\% & 61.0\% & 0.0\% & 22.0\% & 0.0\% & 31.3\% & 0.0\% \\
\textbf{VPP} & 15.0\% & 12.0\% & \underline{40.0\%} & 4.0\% & 59.0\% & 3.0\% & 15.0\% & 0.0\% & 35.8\% & 4.0\% \\
\textbf{RDT} & 33.0\% & \underline{18.0\%} & 35.0\% & 7.0\% & 74.0\% & 27.0\% & 27.0\% & 4.0\% & 47.4\% & \underline{15.1\%} \\
{$\mathbf{\pi_0}$} & \textbf{68.0\%} & \underline{18.0\%} & 28.0\% & 6.0\% & \textbf{91.0\%} & \underline{41.0\%} & \textbf{54.0\%} & \textbf{13.0\%} & \underline{57.1\%} & 14.1\% \\
{$\mathbf{\pi_{0.5}}$} & 46.0\% & 15.0\% & 28.0\% & \underline{12.0\%} & \underline{85.0\%} & \textbf{44.0\%} & 25.0\% & 5.0\% & 45.4\% & 13.3\% \\ 
\textbf{Ours} & \underline{52.0\%} & \textbf{37.0\%} & \textbf{41.0\%} & \textbf{18.0\%} & 80.0\% & 36.0\% & \underline{37.0\%} & \underline{7.0\%} & \textbf{57.5\%} & \textbf{25.5\%} \\
\bottomrule
\end{tabular*}
\end{table*}

As presented in \Cref{tab:sim_tasks}, under the Easy setting, our method achieves the highest average success rates across all the tasks. This demonstrates that our proposed training paradigm yields substantial improvements over the baseline RDT, effectively expanding its capability boundaries. 

Under the Hard setting, all compared methods only reach very low success rates, because varying distractors, background textures, lighting conditions, and table heights pose huge challenges to models' visual generalization.
However, our model surpasses the most effective $\pi_{0.5}$ by a clear margin.
This indicates that our model better learns the low-level dynamics behind diverse visual observations, thus producing correct actions rather than relying on spurious correlations~\citep{xing2025shortcut}.

\subsection{Comparison of Training Paradigm}
\label{subsec:comparison_paradigm}

In this section, we conduct experiments on various training paradigms under the condition of a fixed total training step count of 20,000. For these experiments, we selected two representative tasks from above: Stack Bowls Two and Pick Dual Bottles. Based on the experimental results in \Cref{tab:paradigm_detailed}, we present the following insights regarding effective training paradigms.

\textbf{Mid-Training is necessary before post-training. }As shown in Paradigms 3 and 4, training solely in the Post-Training stage directly on the base model yields results even lower than the RDT baseline. Similar to how fine-tuning struggles to improve task performance on out-of-distribution tasks in language models~\citep{gururangan2020dontstoppretrainingadapt}, we attribute this phenomenon to the significant distribution mismatch between the RDT pre-training and the introduction of the future frame prediction objective. 
Compared to fine-tuning that only uses generating action chunks as the supervised objective, the single-stage mid-training (Paradigm 1) achieved an average improvement of 4.6\% in overall success rate under the same 20k training steps. Furthermore, a comparison with Paradigm 2 indicates that during the mid-training stage, full-parameter training is required to bridge the substantial distribution gap.

\textbf{Post-Training can enhance model performance in a parameter-efficient manner. }
A comparative analysis of Paradigms 1, 5, and 6 in \Cref{tab:paradigm_detailed} reveals that introducing multi-prefix alone results in a minor performance degradation, whereas combining it with LoRA fine-tuning yields a substantial improvement. This discrepancy arises because scaling up the prefix tuning involves switching to different future teacher supervision models, which possess distinct representation spaces. However, the RDT backbone remains frozen, retaining the alignment with the Theia teacher model acquired during the mid-training phase. This shift in external supervision, without a corresponding mechanism for model adaptation, negatively impacts performance. 
Furthermore, after introducing LoRA fine-tuning, the model demonstrated a significant improvement in task success rate compared to both the baseline and the mid-trained model. This indicates the effectiveness of our parallel scaling approach. Moreover, unlike the performance observed during the mid-training stage, the post-training stage does not require full-parameter fine-tuning; only LoRA fine-tuning is sufficient to adapt to the representation space gap caused by changes in the teacher model.

\begin{table*}[t]
\centering
\caption{\textbf{Comparison of training paradigms.}
Please note that \textbf{Bold} and \underline{Underline} denote the best and the second-best performances, respectively, across the paradigms.
The results indicate that both mid-training and post-training phases are significant for the method's improvement.}
\label{tab:paradigm_detailed}
\resizebox{0.8\textwidth}{!}{%
    \begin{tabular}{c l c c c c} 
    \toprule
    Paradigm & \multirow{2}{*}{Method} & \multirow{2}{*}{Training Steps} & \multicolumn{3}{c}{Success Rates (\%)} \\ 
    \cmidrule(lr){4-6}
    No. & & & Easy & Hard & Average \\ 
    \midrule
    0 & RDT   & 20k & 59.0 & 20.5 & 39.8 \\
    1 & mid-train (full ft) & 20k & 63.0   & \underline{27.5} & \underline{45.3} \\
    2 & mid-train (prefix \& lora ft) & 20k & 48.0 & 8.5 & 28.3 \\
    3 & post-train (prefix ft) & 20k & 25.0 & 4.0  & 14.5 \\
    4 & post-train (prefix \& lora ft) & 20k  & 46.0   & 9.0  & 27.5 \\
    5 & mid-train (full ft) + post-train (prefix ft) & 15k + 5k & \underline{68.0} & 21.5 & 44.8 \\
    6 & mid-train (full ft) + post-train (prefix \& lora ft) \textbf{(ours)} & 15k + 5k & \textbf{73.5} & \textbf{32.0} & \textbf{52.3} \\
    \bottomrule
    \end{tabular}%
}
\end{table*}

\begin{table}[t!]
  \centering
  \caption{\textbf{Comparison of Inference Efficiency}}
  \label{tab:efficiency_comparison}
  \resizebox{0.5\columnwidth}{!}{
    \begin{tabular}{lcccc} 
      \toprule
        
        & \makecell[t]{RDT \\ (5 steps)} & \makecell[t]{mid-train \\ (5 steps)} & \makecell[t]{post-train \\ (5 steps)} & \makecell[t]{post-train \\ (3 steps)}  \\  
      \midrule
      Inference Memory (GB) & 3.7     & 3.7       & 8.0     & 8.0 \\
      Latency (Sec)         & 0.214   & 0.228     & 0.235   & 0.173 \\
      Success Rates (\%)         & 39.8   & 45.3     &  52.3  & 48.5 \\
      \bottomrule
    \end{tabular}
  }
\end{table}

\subsection{Efficiency during Training and Inference}
In addition to the training paradigm, we also examined the inference efficiency, a particular concern in the robotics domain. Since our parallel scaling scheme introduces additional parallel computations and parameters, we report a comparison in \Cref{tab:efficiency_comparison} with the vanilla RDT-1B and the mid-trained model in terms of GPU memory usage, inference latency, and average performance on RoboTwin. 

We conducted inference testing using a single NVIDIA H100 GPU. By leveraging CUDA Graph, we have significantly improved inference efficiency. When denoising with the same 5 steps as the baseline model, our action generation latency increased by only about 20 ms. The memory usage has increased to 8.0 GB, yet it remains within the typical memory capacity of commonly used inference GPUs. When the denoising steps were reduced to 3, our method achieved higher performance than the baseline with lower inference latency.

\subsection{Verification Experiments on Smaller-Scale Policy Model}
\label{subsec:comparison_scale}

\begin{figure*}[t]  
    \centering
    \includegraphics[width=0.99\linewidth]{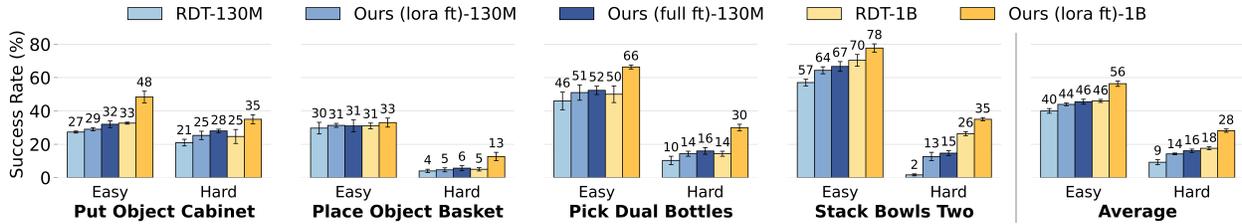} 
    \caption{\textbf{Experiments on different parameter scales on 4 tasks on the RoboTwin 2.0. } The 130M backbone model fine-tuned using either LoRA or full-parameter fine-tuning under the \method~consistently outperforms the naively fine-tuned RDT-130M across all tasks and remains competitive when compared with the naively fine-tuned RDT-1B. Especially in the \textbf{Stack Bowls Two-Hard} task, the improvement is significant.}
    \label{fig:paradigms} 
\end{figure*}

We further validated the proposed paradigm on a model with significantly smaller parameters (RDT-130M). As shown in \Cref{fig:paradigms}, our paradigm continues to demonstrate outstanding performance even on a smaller baseline below the 1B scale.

First, compared to the baseline model, our method achieves consistent performance improvements across both easy and hard task settings. Second, at the 130M parameter scale, the performance of LoRA fine-tuning during the post-training stage is only 2–3\% lower in average success rate across various tasks compared to full-parameter fine-tuning. This strongly demonstrates the high efficiency of applying LoRA fine-tuning within our architectural framework.

Moreover, our optimized small model exhibits performance comparable to the RDT-1B baseline. Through further analysis of individual subtasks, we observed that RDT-130M shows weaker generalization on hard-type tasks compared to the 1B baseline. However, our training paradigm delivers substantial improvements on all hard tasks. This provides compelling evidence that our training scheme and architectural design substantially extend the capability boundaries of the model.

\begin{figure*}[t]  
    \centering
    \includegraphics[width=0.9\linewidth]{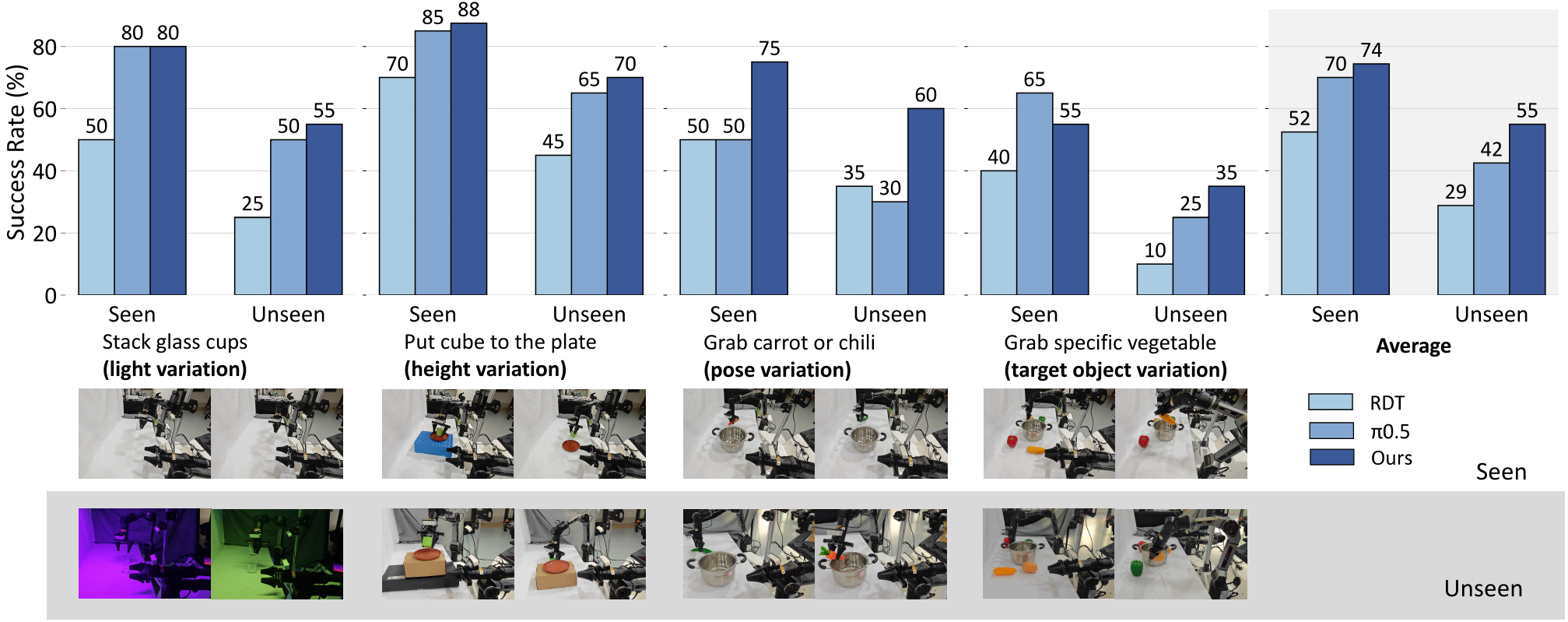} 
    \caption{\textbf{Real-world experiment results in seen and unseen scenarios.} We evaluate our \method~and prior SOTA VLAs on 4 representative tasks, each with different axes of generalization. Seen means the settings were included in the training data, while Unseen refers to new task settings that the model did not encounter during training.}
    \label{fig:easycobot} 
\end{figure*}

\section{Real-world Experiments}
In real-world experiments, we focus on exploring the model's capabilities in more complex scenarios (\Cref{subsec:comparison_complex}) and the scalability of the training scheme at the data level (\Cref{subsec:co_training}).

\subsection{Experimental Setup} 

Real-world experiments are conducted on a bimanual AgileX mobile manipulator. Each arm features six degrees of freedom (6-DoF) and is equipped with a parallel gripper. The vision system comprises a primary high-mounted camera for a third-person perspective and two wrist-mounted cameras for ego-centric views. For training, we utilize 25 demonstrations per variation for basic tasks, and 100 demonstrations for long-horizon tasks. During evaluation, we perform 40 trials for each basic task and 20 trials for each long-horizon task.

\subsection{Comparison on Generalization and Long-horizon Tasks}
\label{subsec:comparison_complex}

We designed four basic tasks to test our model under different variations: lighting, height, pose, and object variations. Each task is divided into seen and unseen settings to evaluate the vision generalization capacity of the fine-tuned model. As shown in \Cref{fig:easycobot}, our model achieves strong performance in all cases, especially in unseen ones. These results match our findings in the simulation and show that our model has good generalization in the real world.

\begin{figure}[t]  
    \centering
    \includegraphics[width=0.6\linewidth]{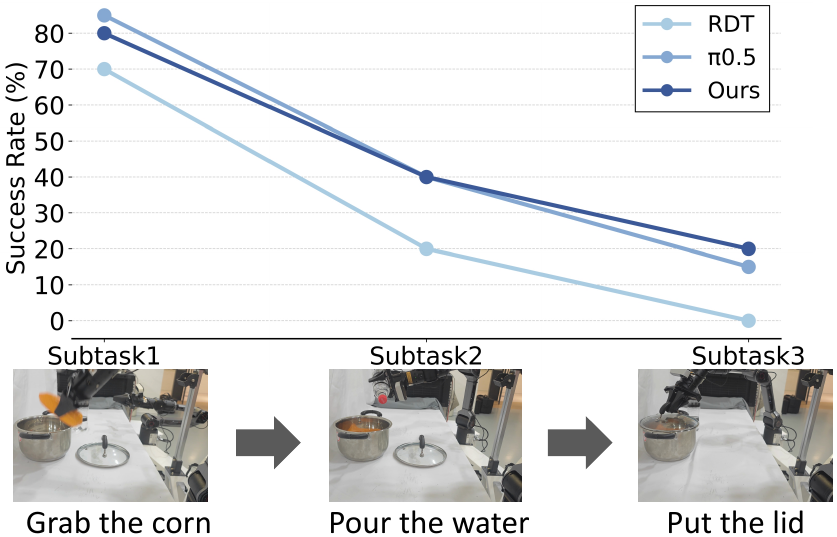} 
    \caption{\textbf{Long-horizon Performance.} Each data point represents the success rate of completing up to and including that corresponding subtask.}
    \label{fig:longcobot} 
\end{figure}

To investigate the performance limits of \method, we devised a challenging long-horizon task characterized by three temporally dependent sub-tasks and four interactive objects. This task necessitates sophisticated dual-arm coordination, as illustrated in \Cref{fig:longcobot}. Our results demonstrate that \method~significantly outperforms the baseline RDT. Specifically, while the vanilla RDT failed to complete the task in any trial due to challenges in fine-grained manipulation, such as precisely grasping and placing a lid, and maintaining action continuity across sub-tasks. In contrast, our model achieved a 20\% success rate.

\subsection{Co-Training with Human Egocentric Data}
\label{subsec:co_training}
In this section, we explain how the model benefits from action-free human egocentric data.
We demonstrate that \method~can leverage such data to enhance the generalized world-modeling capabilities of the base model, rather than being effective only in fine-tuning processes that involve robot data.

\paragraph{Data Pyramid}
Motivated by these insights, we propose a hierarchical data pyramid framework to strategically integrate diverse data types across training stages.

The \textbf{bottom} layer of our pyramid consists of large-scale action-free human egocentric data, which are usually open-sourced datasets with hundreds to thousands of hours of recordings and include a wide variety of contact-rich manipulation tasks~\citep{TASTE-Rob}.
The \textbf{middle} layer comprises task-specific human egocentric data, collected by human operators directly manipulating objects without teleoperating a robotic arm. This makes data collection highly efficient and requires only minimal hardware. Since pretrained VLA models assume a fixed third-person camera setting, we do not use head-mounted cameras such as GoPro or VR devices~\citep{flare}. Instead, we adopt a static third-person camera setup consistent with conventional robot data collection.
Under this setup, even inexperienced human operators can achieve a data collection rate of more than 360 trajectories per hour.
The \textbf{top} layer of the pyramid consists of conventional task-specific robot data, collected via human teleoperation of the robot using dedicated control interfaces. Skilled operators can typically collect up to 120 trajectories per hour in this setting.

\paragraph{Co-Training}
We train the model during the mid-training phase by mixing web-scale human egocentric data with task-specific human data and robot data for sampling.
When training on action-free samples, the action loss in \Cref{eq:rdt_loss} is omitted and instead only the alignment loss is optimized. 

To evaluate whether our method can effectively learn from human egocentric videos and further scale with large amounts of task-irrelevant egocentric data, we compare the performance of models trained under different data configurations on a pick-and-place task. The target objects consist of five easy-to-grasp objects and five hasy-to-grasp objects.
The training data used in this experiment contains 5 robot action trajectories per object, 50 task-specific human egocentric trajectories, and 10k task-irrelevant human egocentric videos. 

Here, we consider four training settings:
\begin{itemize}
    \item \textbf{Robot (task):} Fine-tuning on robot data only.
    
    \item \textbf{+ Ego (task):} Fine-tuning on a mixture of robot and human egocentric data.
    
    \item \textbf{+ Ego (web):} Fine-tuning on both web-scale human egocentric data and robot data. Detailed information on Ego (web) can be found in Appendix.
    
    \item \textbf{+ Ego (task) + Ego (web):} Co-training with all three types of data.
\end{itemize}

\begin{figure*}[t]  
    \centering
    \includegraphics[width=0.99\linewidth]{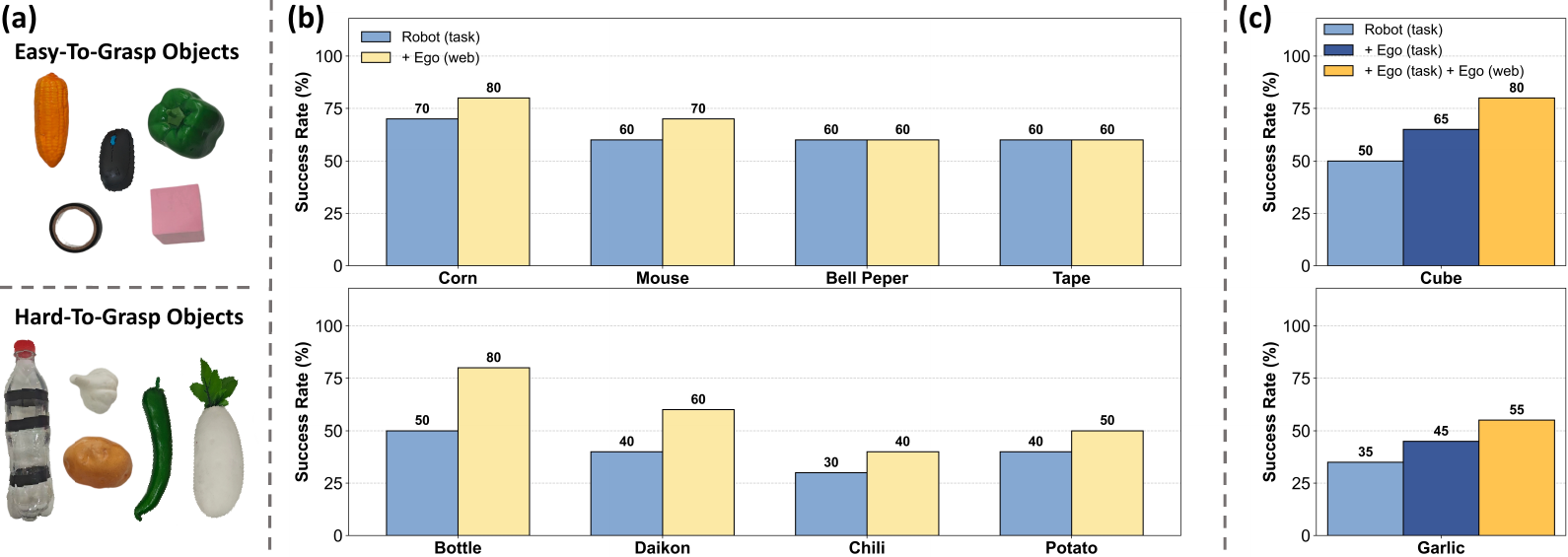} 
    \caption{\textbf{Leveraging human egocentric data without action labels.} \textbf{(a).} Experimental targets, including 5 objects that are easy to pick up, and 5 objects that are difficult to pick up. 
    \textbf{(b).} Multi-task results of \method~trained on \textbf{Robot (task)} and \textbf{+ Ego (web)} across 8 target object categories. Here, ``multi-task'' denotes that one model is used to be trained and tested on multiple target objects. \textbf{(c).} The results of \method~trained on \textbf{Robot (task)}, \textbf{+ Ego (web)} and \textbf{+ Ego (task) + Ego (web)} data mixture, respecitvely.}
    \label{fig:ego_exp} 
\end{figure*}

Based on the results, the following conclusions can be drawn:

\textbf{Training on large-scale open-sourced egocentric videos provides a strong inductive prior for the model to handle novel objects.}
\Cref{fig:ego_exp}(b) shows that for the 5 easy-to-grasp objects, models trained on large-scale egocentric data consistently achieve performance improvements. This suggests that although the open-sourced egocentric dataset is irrelevant to our downstream tasks, training on it enhances the model’s world modeling capabilities.

For the 5 hard-to-grasp objects, models trained with only a small amount of robot data exhibit low success rates, whereas models with co-training achieve substantially higher performance. 
We believe this is likely because the large-scale dataset naturally offers a diverse distribution of complex object geometries, thereby providing a stronger inductive prior for the model.

\textbf{\method~can leverage task-specific egocentric data to improve spatial generalization.} 
We select two representative target objects (cube and garlic) and observe that training on only 5 robot trajectories results in very low success rates, as shown in \Cref{fig:ego_exp}(c). This indicates that although \method~exhibits few-shot learning capability, the extremely limited number of samples still prevents the model from learning the task dynamics and thus from achieving position generalization.
On the other hand, incorporating human egocentric data enables the model to learn future prediction and capture underlying dynamics from more samples, resulting in significant improvements.

\textbf{Combining data from different levels of the data pyramid enables effective scaling and leads to maximized performance.} Furthermore, the results in \Cref{fig:ego_exp}(c) show that when the advantages of training are combined with co-training samples, model performance is further improved. 

\section{Conclusion} 
\label{sec:conclusion}
This paper proposes~\method, an efficient training scheme built upon pre-trained diffusion-based VLA models. It includes a mid-training phase for world-modeling enhanced visual representation alignment and a parameter-efficient post-training phase to unlock the parallel scaling capacity of the model.

Extensive simulation experiments on the RoboTwin platform and real-world evaluations demonstrate that \method~exhibits significant advantages in data efficiency, parameter efficiency, and overall performance compared to baseline models trained through direct fine-tuning. Furthermore, our model can effectively leverage large-scale egocentric human operation data, which further enhances the generalization capability of the pre-trained model beyond merely fine-tuning on target domains.


\bibliographystyle{assets/plainnat}
\bibliography{paper}

\clearpage
\newpage
\onecolumn
\beginappendix
\renewcommand{\thefigure}{S\arabic{figure}}
\renewcommand{\thetable}{S\arabic{table}}
\setcounter{figure}{0}
\setcounter{table}{0}

In this appendix, we provide supplementary information regarding the implementation details and hyperparameter analysis of our proposed framework. \Cref{app:hypers} presents the empirical studies used to determine the optimal configuration for the alignment loss coefficient, future supervision depth, and prediction horizon. \Cref{app:ego_co} details the human egocentric co-training phase, including dataset characteristics, computational requirements, and the rationale for our data selection.  
\section{Hyperparameter Settings}
\label{app:hypers}

\begin{table}[ht]
\centering
\caption{weight factor of alignment loss.}
    \begin{tabular}{c|cccccc}
    \toprule
    $\lambda_1$ & 0 & 0.001 & 0.02 & 0.05 & 0.1 & 0.5  \\
    \midrule
    SR (\%) & 14.0 & 18.5 & 26.4 & \textbf{32.5} & 22.0 & 23.5 \\
    \bottomrule
\end{tabular}
\label{tab:alpha}
\end{table}

The hyper-parameter $\lambda_1$ serves as the coefficient for the alignment loss in \Cref{eq:total_mipa_loss}. An appropriate value for this coefficient enables the model to effectively learn future state prediction and capture the dynamics between actions and states. However, an excessively large weight may interfere with the model's primary task of action prediction.
As shown in \Cref{tab:alpha}, the model performs best under the $\lambda_1=0.05$.

\begin{table}[ht]
\centering
\caption{depth of future alignment.}
    \begin{tabular}{c|cccc}
    \toprule
    $depth$ & 7 & 14 & 21 & 28   \\
    \midrule
    SR (\%) & 14.5 & 18.0 & \textbf{23.5} & 16.0 \\
    \bottomrule
\end{tabular}
\label{tab:depth}
\end{table}

A critical design choice in \Cref{eq:align_loss_S} involves determining the optimal depth of $\mathbf{p}_t$ within the DiT layers of RDT to facilitate effective future supervision. The RDT-1B backbone comprises a total of 28 DiT layers. As shown in \Cref{tab:depth}, we find that utilizing the output of the 21st layer for learnable prefix alignment yields the best performance. This observation is consistent with the findings in~\citep{flare}, suggesting that supervising the representations at approximately three-quarters of the total depth of the model is the most beneficial.

\begin{table}[ht]
\centering
\caption{steps of future alignment.}
    \begin{tabular}{c|cccc}
    \toprule
    $h$ & 8 & 16 & 32  \\
    \midrule
    SR (\%) & \textbf{35.3} & 35.0 & 29.7 \\
    \bottomrule
\end{tabular}
\label{tab:h}
\end{table}

The selection of the future horizon $h$ significantly influences the ability of the model to capture future representations. Our experimental results, as summarized in \Cref{tab:h}, demonstrate that the model achieves optimal performance when $h$ is set to 8.

Based on these empirical findings, we use the optimal parameters identified above as the standard configuration for all other experiments.

\section{Human Egocentric Co-Training Details}
\label{app:ego_co}

We provide a comprehensive overview of the training stage involving web-
scale human egocentric data. We leverage TASTE-Rob \citep{TASTE-Rob}, a large-scale dataset of ego-centric hand-object interaction videos containing 100,856 video sequences and roughly 9 million frames, with high-fidelity linguistic alignment for each video. The training involves one epoch and requires 96 hours of computation on 8 H100 GPUs. The rationale for utilizing TASTE-Rob lies in its fixed egocentric viewpoint, which closely mirrors the camera settings typically found in mainstream VLA models. Such consistency significantly enhances the transferability of learned features for downstream robot action prediction tasks.

\end{document}